%% file: ijcai24.tex
\documentclass{article}
\usepackage{ijcai24}

\usepackage{times}
\usepackage{microtype}
\usepackage{soul}
\usepackage{url}
\usepackage[usenames,svgnames]{xcolor}
\usepackage[colorlinks,
citecolor=NavyBlue,
linkcolor=NavyBlue,
urlcolor=NavyBlue]{hyperref}
\usepackage[utf8]{inputenc}
\usepackage{setspace}
\usepackage[small]{caption}
\usepackage{graphicx}
\usepackage{amsmath}
\usepackage{amsthm}
\usepackage{booktabs}
\usepackage{algorithm}
\usepackage{algorithmic}
\usepackage{mydef}
\usepackage{booktabs}
\usepackage{tabularx}
\usepackage{framed}
\usepackage{bbding}
\usepackage{subfloat}
\usepackage{adjustbox}
\usepackage{multirow}
\usepackage{pifont}
\usepackage{natbib}

\newcommand{\paratitle}[1]{\vspace{0.8ex}\noindent \textbf{#1}}

\newcommand{\figref}[1]{Figure~\ref{#1}}
\newcommand{\tabref}[1]{Table~\ref{#1}}
\newcommand{\secref}[1]{Section~\ref{#1}}

\title{A Survey of Graph Meets Large Language Model: Progress and Future Directions}

\author{
Yuhan Li$^1$\footnote{Equal Contribution}\and
Zhixun Li$^2$\footnotemark[1]\and
Peisong Wang$^{3}$\footnotemark[1]\and
Jia Li$^1$\footnote{Corresponding author (\href{jialee@ust.hk}{jialee@ust.hk}).}\and
Xiangguo Sun$^2$\and \\
Hong Cheng$^2$\and 
Jeffrey Xu Yu$^2$
\affiliations
$^1$The Hong Kong University of Science and Technology (Guangzhou)\\
$^2$The Chinese University of Hong Kong\\
$^3$Tsinghua University\\
}

\begin{document}

\maketitle

\input{1_abs}
\input{2_intro}

\input{3_pre}
\input{4_enhancer}
\input{5_predictor}

\input{6_align}

\input{7_future}
\input{8_con}

\bibliographystyle{named}
\bibliography{mybib}

\end{document}

%% file: 1_abs.tex
\begin{abstract}
    Graph plays a significant role in representing and analyzing complex relationships in real-world applications such as citation networks, social networks, and biological data. Recently, Large Language Models (LLMs), which have achieved tremendous success in various domains, have also been leveraged in graph-related tasks to surpass traditional Graph Neural Networks (GNNs) based methods and yield state-of-the-art performance. In this survey, we first present a comprehensive review and analysis of existing methods that integrate LLMs with graphs. First of all, we propose a new taxonomy, which organizes existing methods into three categories based on the role (i.e., enhancer, predictor, and alignment component) played by LLMs in graph-related tasks. Then we systematically survey the representative methods along the three categories of the taxonomy. Finally, we discuss the remaining limitations of existing studies and highlight promising avenues for future research. The relevant papers are summarized and will be consistently updated at: \url{https://github.com/yhLeeee/Awesome-LLMs-in-Graph-tasks}.
\end{abstract}

%% file: 2_intro.tex
\section{Introduction}

Graph, or graph theory, serves as a fundamental part of numerous areas in the modern world, particularly in technology, science, and logistics \citep{ji2021survey}. Graph data represents the structural characteristics between nodes, thus illuminating relationships within the graph's components. Many real-world datasets, such as citation networks \citep{sen2008collective}, social networks \citep{hamilton2017inductive}, and molecular \citep{wu2018moleculenet}, are intrinsically represented as graphs. To tackle graph-related tasks, Graph Neural Networks (GNNs) \citep{kipf2016semi,velickovic2018graph} have emerged as one of the most popular choices for processing and analyzing graph data. The main objective of GNNs is to acquire expressive representations at the node, edge, or graph level for different kinds of downstream tasks through recursive message passing and aggregation mechanisms among nodes.

In recent years, significant advancements have been made in Large Language Models (LLMs) like Transformers \citep{vaswani2017attention}, BERT \citep{kenton2019bert}, GPT \citep{brown2020language}, and their variants. These LLMs can be easily applied to various downstream tasks with little adaptation, demonstrating remarkable performance across various natural language processing tasks, such as sentiment analysis, machine translation, and text classification \citep{zhao2023survey}. While their primary focus has been on text sequences, there is a growing interest in enhancing the multi-modal capabilities of LLMs to enable them to handle diverse data types, including graphs \citep{chai2023graphllm}, images \citep{zhang2023llama}, and videos \citep{zhang2023video}. 

\begin{figure}[t]
  \centering
  \includegraphics[width=0.48\textwidth]{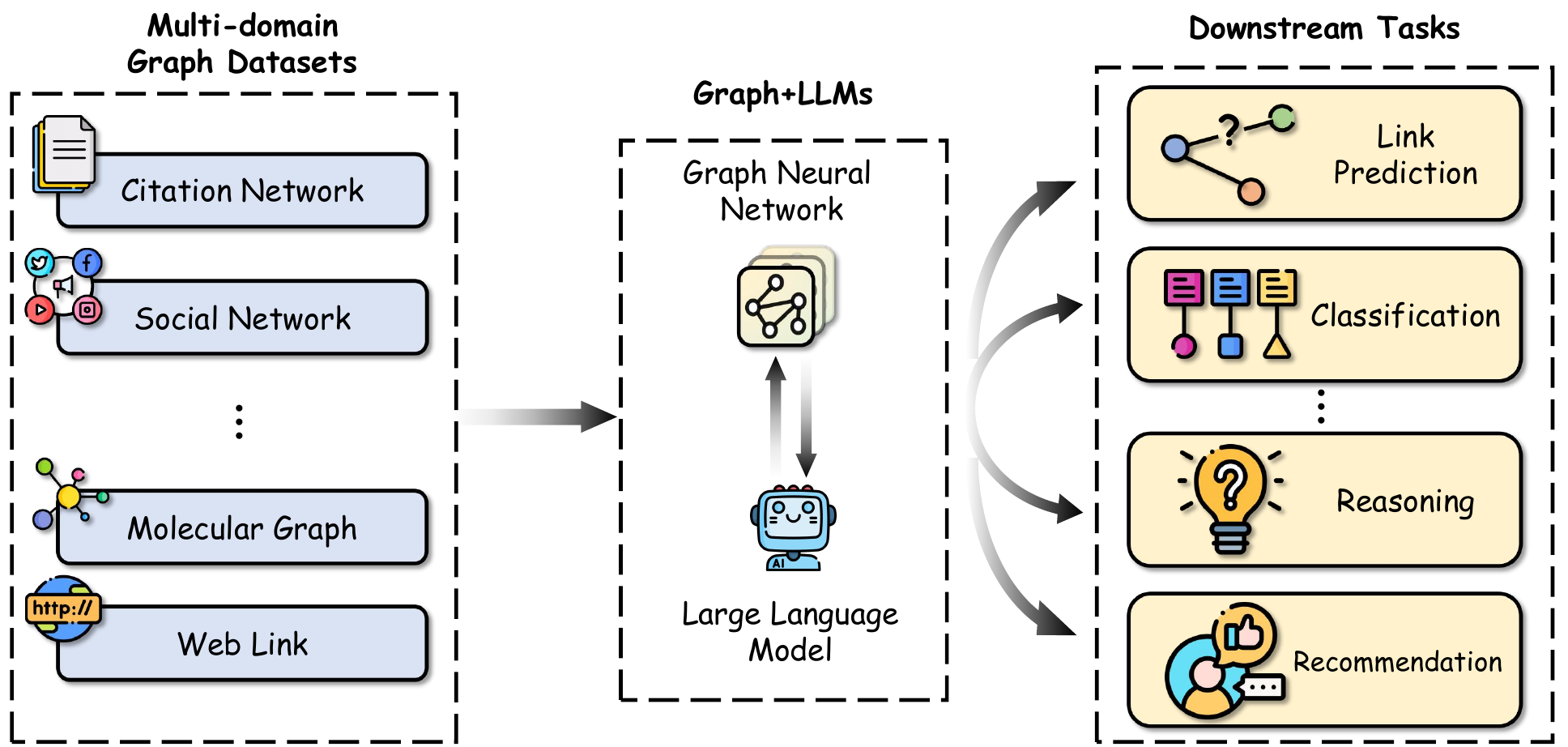}
  \caption{Across a myriad of graph domains, the integration of graphs and LLMs demonstrates success in various downstream tasks.}
  \label{fig:overview}
\end{figure}

\paratitle{LLMs help graph-related tasks.} With the help of LLMs, there has been a notable shift in the way we interact with graphs, particularly those containing nodes associated with text attributes. As shown in Figure \ref{fig:overview}, the integration of graphs and LLMs demonstrates success in various downstream tasks across a myriad of graph domains. Integrating LLMs with traditional GNNs can be mutually beneficial and enhance graph learning. While GNNs are proficient at capturing structural information, they primarily rely on semantically constrained embeddings as node features, limiting their ability to express the full complexities of the nodes. Incorporating LLMs, GNNs can be enhanced with stronger node features that effectively capture both structural and contextual aspects. On the other hand, LLMs excel at encoding text but often struggle to capture structural information present in graph data. Combining GNNs with LLMs can leverage the robust textual understanding of LLMs while harnessing GNNs' ability to capture structural relationships, leading to more comprehensive and powerful graph learning. For example, TAPE \citep{he2023explanations} leverages semantic knowledge that is relevant to the nodes (i.e., papers) generated by LLMs to improve the quality of initial node embeddings in GNNs. In addition, InstructGLM \citep{ye2023natural} replaces the predictor from GNNs with LLMs, leveraging the expressive power of natural language through techniques such as flattening graphs and designing instruction prompts. MoleculeSTM \citep{liu2022multi} aligns GNNs and LLMs into the same vector space to introduce textual knowledge into graphs (i.e., molecules), thereby improving reasoning abilities.

\input{figures/taxonomy}

It is evident that LLMs have a significant influence on graph-related tasks from different perspectives. To achieve a better systematic overview, as shown in \figref{fig:taxonomy_of_pGMs}, we follow \cite{chen2023exploring} to organize our first-level taxonomy, categorizing based on the role (i.e., enhancer, predictor, and alignment component) played by LLMs throughout the entire model pipeline. We further refine our taxonomy and introduce more granularity to the initial categories.

\paratitle{Motivations.} Although LLMs have been increasingly applied in graph-related tasks, this rapidly expanding field still lacks a systematic review. \cite{zhang2023large} conducts a forward-looking survey, presenting a perspective paper that discusses the challenges and opportunities associated with the integration of graphs and LLMs. \cite{liu2023towards} provide another related survey that summarizes existing graph foundation models and offers an overview of pre-training and adaptation strategies. However, both of them have limitations in terms of comprehensive coverage and the absence of a taxonomy specifically focused on how LLMs enhance graphs. In contrast, we concentrate on scenarios where both graph and text modalities coexist and propose a more fine-grained taxonomy to systematically review and summarize the current status of LLMs techniques for graph-related tasks.

\paratitle{Contributions.} The contributions of this work can be summarized from the following three aspects. \textbf{(1)} \textit{A structured taxonomy}. A broad overview of the field is presented with a structured taxonomy that categorizes existing works into four categories (\figref{fig:taxonomy_of_pGMs}). \textbf{(2)} \textit{A comprehensive review}. Based on the proposed taxonomy, the current research progress of LLMs for graph-related tasks is systematically delineated. \textbf{(3)} \textit{Some future directions}. We discuss the remaining limitations of existing works and point out possible future directions.

%% file: figures/taxonomy.tex
\tikzstyle{leaf}=[draw=hiddendraw,
    rounded corners,minimum height=1em,
    fill=mygreen!40,text opacity=1, align=center,
    fill opacity=.5,  text=black,align=left,font=\scriptsize,
    inner xsep=3pt,
    inner ysep=1pt,
    ]
\tikzstyle{middle}=[draw=hiddendraw,
    rounded corners,minimum height=1em,
    fill=output-white!40,text opacity=1, align=center,
    fill opacity=.5,  text=black,align=left,font=\scriptsize,
    inner xsep=3pt,
    inner ysep=1pt,
    ]
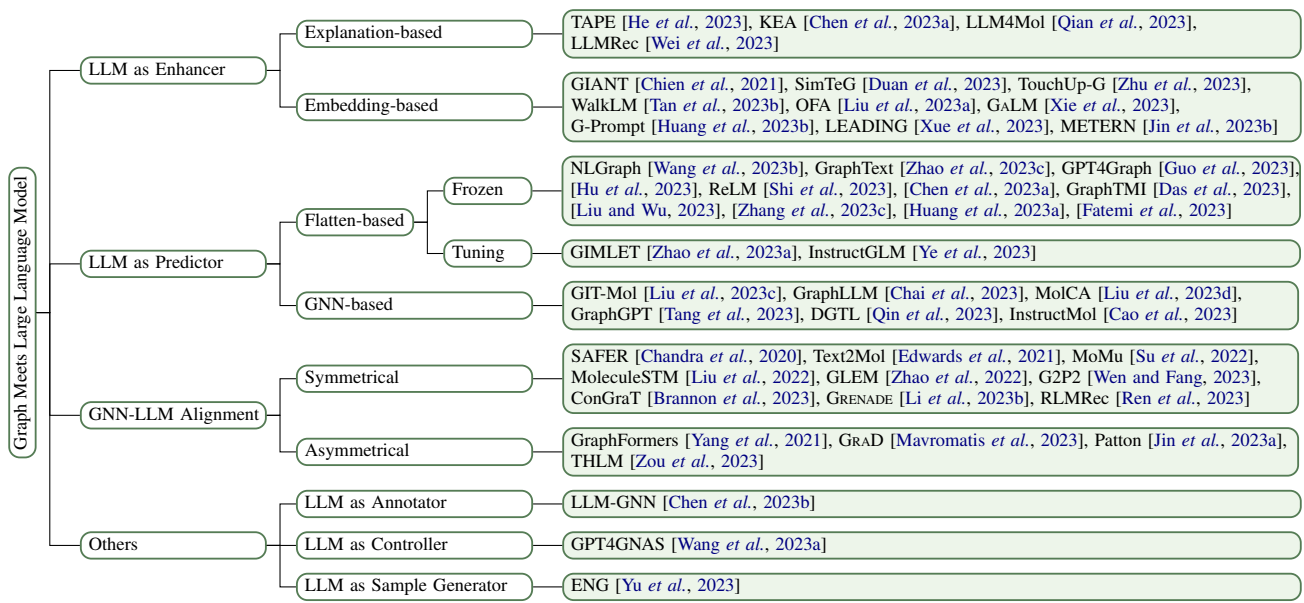
\begin{figure*}[ht]
\centering
\begin{forest}
  for tree={
  forked edges,
  grow=east,
  reversed=true,
  anchor=base west,
  parent anchor=east,
  child anchor=west,
  base=middle,
  font=\scriptsize,
  rectangle,
  line width=0.7pt,
  draw=output-black,
  rounded corners,align=left,
  minimum width=2em,
    s sep=5pt,
    inner xsep=3pt,
    inner ysep=1pt,
  },
  where level=1{text width=4.5em}{},
  where level=2{text width=6em,font=\scriptsize}{},
  where level=3{font=\scriptsize}{},
  where level=4{font=\scriptsize}{},
  where level=5{font=\scriptsize}{},
  [Graph Meets Large Language Model, middle,rotate=90,anchor=north,edge=output-black
    [LLM as Enhancer, middle, edge=output-black,text width=6.4em
        [Explanation-based, middle, text width=8.3em, edge=output-black
            [TAPE \citep{he2023explanations}{,} KEA \citep{chen2023exploring}{,} LLM4Mol \citep{qian2023can}{,}\\ LLMRec \citep{wei2023llmrec}, leaf, text width=27.3em, edge=output-black]
        ]
        [Embedding-based, middle, text width=8.3em, edge=output-black
            [GIANT \citep{chien2021node}{,} SimTeG \citep{duan2023simteg}{,} TouchUp-G \citep{zhu2023touchup}{,}\\ WalkLM \citep{tan2023walklm}{,} OFA \citep{liu2023one}{,} G$\scriptstyle \text{A}$LM \citep{xie2023graph}{,}\\ G-Prompt \citep{huang2023prompt}{,} LEADING \citep{xue2023efficient}{,} METERN \citep{jin2023learning}, leaf, text width=27.3em, edge=output-black]
        ]
    ]
    [LLM as Predictor, middle, edge=output-black, text width=6.4em
        [Flatten-based, middle, text width=3.8em, edge=output-black
            [Frozen, middle, text width=2.7em, edge=output-black
                [NLGraph \citep{wang2023can}{,}  GraphText \citep{zhao2023graphtext}{,} GPT4Graph \citep{guo2023gpt4graph}{,} \\ \citep{hu2023beyond}{,}  ReLM \citep{shi2023relm}{,} \citep{chen2023exploring}{,}  GraphTMI \citep{das2023modality}{,} \\  \citep{liu2023evaluating}{,}   \citep{zhang2023llm4dyg}{,} \citep{huang2023can}{,} \citep{fatemi2023talk},   leaf, text width=27.3em, edge=output-black]
            ]
            [Tuning, middle, text width=2.7em, edge=output-black
                [GIMLET \citep{zhao2023gimlet}{,} InstructGLM \citep{ye2023natural}, leaf, text width=27.3em, edge=output-black]
            ]
        ]
        [GNN-based, middle, text width=8.3em, edge=output-black
            [GIT-Mol  \citep{liu2023git}{,} GraphLLM \citep{chai2023graphllm}{,} MolCA \citep{liu2023molca}{,} \\ GraphGPT \citep{tang2023graphgpt}{,} DGTL \citep{qin2023disentangled}{,} InstructMol \citep{cao2023instructmol}, leaf, text width=27.2em, edge=output-black]
        ]
    ]
    [GNN-LLM Alignment, middle, edge=output-black, text width=6.4em
        [Symmetrical, middle, text width=8.3em, edge=output-black
            [SAFER \citep{chandra2020graph}{,} Text2Mol \citep{edwards2021text2mol}{,} MoMu \citep{su2022molecular}{,} \\MoleculeSTM \citep{liu2022multi}{,} GLEM \citep{zhao2022learning}{,} G2P2 \citep{wen2023prompt}{,}  \\ConGraT \citep{brannon2023congrat}{,} G$\scriptstyle \text{RENADE}$ \citep{li2023grenade}{,} RLMRec \citep{ren2023representation}, leaf, text width=27.2em, edge=output-black]
        ]
        [Asymmetrical, middle, text width=8.3em, edge=output-black
            [GraphFormers \citep{yang2021graphformers}{,}  G$\scriptstyle \text{RA}$D \citep{mavromatis2023train}{,} Patton \citep{jin2023patton}{,}\\THLM \citep{zou2023pretraining},leaf, text width= 27.2em, edge=output-black]
        ]
    ]
    [Others, middle, edge=output-black, text width=6.4em
        [LLM as Annotator, middle, edge=output-black, text width=8.3em
            [LLM-GNN \citep{chen2023label}, leaf, text width=27.3em, edge=output-black]
        ]
        [LLM as Controller, middle, edge=output-black, text width=8.3em
            [GPT4GNAS \citep{wang2023graph}, leaf, text width=27.3em, edge=output-black]
        ]
        [LLM as Sample Generator, middle,
        % Simulator
        edge=output-black, text width=8.3em
            [ENG \citep{yu2023empower}, leaf, text width=27.3em, edge=output-black]
        ]
     ]
  ]
\end{forest} % Love NiNi 
\caption{A taxonomy of models for solving graph tasks with the help of large language models (LLMs) with representative examples.}
\label{fig:taxonomy_of_pGMs}
\end{figure*}

%% file: 3_pre.tex
\section{Preliminary}

In this section, we first introduce the basic concepts of two key areas related to this survey, i.e., GNNs and LLMs. Next, we give a brief introduction to the newly proposed taxonomy.

\subsection{Graph Neural Networks}

\paratitle{Definitions.} Most existing GNNs follow the message-passing paradigm which contains message aggregation and feature update, such as GCN \citep{kipf2016semi} and GAT \citep{velickovic2018graph}. They generate node representations by iteratively aggregating information of neighbors and updating them with non-linear functions. The forward process can be defined as:
\begin{equation}
    h_i^{(l)}=\mathbf{U}\Big(h_i^{(l-1)}, \mathbf{M}(\{h_i^{(l-1)}, h_j^{(l-1)}|v_j\in\mathcal{N}_i\})\Big)
    \nonumber
\end{equation}
where $h_i^{(l)}$ is the feature vector of node $i$ in the $l$-th layer, and $\mathcal{N}_i$ is a set of neighbor nodes of node $i$. $\mathbf{M}$ denotes the message passing function of aggregating neighbor information, $\mathbf{U}$ denotes the update function with central node feature and neighbor node features as input. By stacking multiple layers, GNNs can aggregate messages from higher-order neighbors.

\paratitle{Graph pre-training and prompting.} While GNNs have achieved some success in graph machine learning, they require expensive annotations and barely generalize to unseen data. To remedy these deficiencies, graph pre-training aims to extract some general knowledge for the graph models to easily deal with different tasks without significant annotation cost. The current mainstream graph pertaining methods can be divided into contrastive and generative approaches. For instance, GraphCL \citep{you2020graph} and GCA \citep{zhu2021graph} follow a contrastive learning framework and maximize the agreement between two augmented views. \cite{sun2023self} extend the contrastive idea to hypergraphs. GraphMAE \citep{hou2022graphmae}, S2GAE \citep{tan2023s2gae}, and WGDN \citep{DBLP:conf/aaai/Cheng0LT23} mask the component of the graph and attempt to reconstruct the original data. The typical learning scheme of ``pre-training and fine-tuning'' is based on the assumption that the pre-training task and downstream tasks share some common intrinsic task space. Instead, in the NLP area, researchers gradually focus on a new paradigm of ``pre-training, prompting, and fine-tuning'', which aims to reformulate input data to fit the pretext. This idea has also been naturally applied to the graph learning area. GPPT \citep{sun2022gppt} first pre-trains graph model by masked edge prediction, then modify the standalone node into a token pair and reformulate the downstream classification as edge prediction task. Additionally, All in One \citep{sun2023all} proposes a multi-task prompting framework, which unifies the format of graph prompts and language prompts.

\subsection{Large Language Models}

\paratitle{Definitions.} While there is currently no clear definition for LLMs \citep{shayegani2023survey}, here we provide a specific definition for LLMs mentioned in this survey. Two influential surveys on LLMs \citep{zhao2023survey,yang2023harnessing} distinguish between LLMs and pre-trained language models (PLMs) from the perspectives of model size and training approach. To be specific, LLMs are those huge language models (i.e., billion-level) that undergo pre-training on a significant amount of data, whereas PLMs refer to those early pre-trained models with moderate parameter sizes (i.e., million-level), which can be easily further fine-tuned on task-specific data to achieve better results to downstream tasks. Due to the relatively smaller parameter size of GNNs, incorporating GNNs and LLMs often does not require LLMs with large parameters. Hence, we follow \cite{liu2023towards} to extend the definition of LLMs in this survey to encompass both LLMs and PLMs as defined in previous surveys. 

\paratitle{Evolution.} LLMs can be divided into two categories based on non-autoregressive and autoregressive language modeling. Non-autoregressive LLMs typically concentrate on natural language understanding and employ a ``masked language modeling'' pre-training task, while autoregressive LLMs focus more on natural language generation, frequently leveraging the ``next token prediction'' objective as their foundational task. Classic encoder-only models such as BERT \citep{kenton2019bert}, SciBERT \citep{beltagy2019scibert}, and RoBERTa \citep{liu2019roberta} fall under the category of non-autoregressive LLMs. Recently, autoregressive LLMs have witnessed continuous development. Examples include Flan-T5 \citep{chung2022scaling} and ChatGLM \citep{zeng2022glm}, which are built upon the encoder-decoder structure, as well as GPT-3 \citep{brown2020language}, PaLM \citep{chowdhery2022palm}, Galactica \citep{taylor2022galactica}, and LLaMA \citep{touvron2023llama}, which are based on decoder-only architectures. Significantly, advancements in architectures and training methodologies of LLMs have given rise to emergent capabilities \citep{wei2022emergent}, which is the ability to handle complex tasks in few-shot or zero-shot scenarios via some techniques such as in-context learning \citep{radford2021learning,dong2022survey} and chain-of-thought \citep{wei2022chain}.

\subsection{Proposed Taxonomy}

We propose a taxonomy (as illustrated in \figref{fig:taxonomy_of_pGMs}) that organizes representative techniques involving both graph and text modalities into three main categories: \textbf{(1)} LLM as Enhancer, where LLMs are used to enhance the classification performance of GNNs. \textbf{(2)} LLM as Predictor, where LLMs utilize the input graph structure information to make predictions. \textbf{(3)} GNN-LLM Alignment, where LLMs semantically enhance GNNs through alignment techniques. We note that in some models, due to the rarity of LLMs' involvement, it becomes difficult to categorize them into these three main classes. Therefore, we separately organize them into the ``Others'' category and provide their specific roles in \figref{fig:taxonomy_of_pGMs}. For example, LLM-GNN \citep{chen2023label} actively selects nodes for ChatGPT to annotate, thereby augmenting the GNN training by utilizing the LLM as an \textit{annotator}. GPT4GNAS \citep{wang2023graph} considers the LLM as an experienced \textit{controller} in the task of graph neural architecture search. It utilizes GPT-4 \citep{openai2023gpt4} to explore the search space and generate novel GNN architectures. Furthermore, ENG \citep{yu2023empower} empowers the LLM as a \textit{sample generator} to generate additional training samples with labels to provide sufficient supervision signals for GNNs.  

In the following sections, we present a comprehensive survey along the three main categories of our taxonomy for incorporating LLMs into graph-related tasks, respectively.

%% file: 4_enhancer.tex
\section{LLM as Enhancer}
\label{sec:enhancer}

GNNs have become powerful tools to analyze graph-structure data. However, the most mainstream benchmark datasets (e.g., Cora \citep{yang2016revisiting} and Ogbn-Arxiv \citep{hu2020open}) adopt naive methods to encode text information in TAGs using shallow embeddings, such as bag-of-words, skip-gram \citep{mikolov2013distributed}, or TF-IDF \citep{salton1988term}. This inevitably constrains the performance of GNNs on TAGs. LLM-as-enhancer approaches correspond to enhancing the quality of node embeddings with the help of powerful LLMs. The derived embeddings are attached to the graph structure to be utilized by any GNNs or directly inputted into downstream classifiers for various tasks. We naturally categorize these approaches into two branches: explanation-based and embedding-based, depending on whether they use LLMs to produce additional textual information.

\subsection{Explanation-based Enhancement}

\begin{figure}[t]
    \centering
    \resizebox{0.9\linewidth}{!}{
        \includegraphics{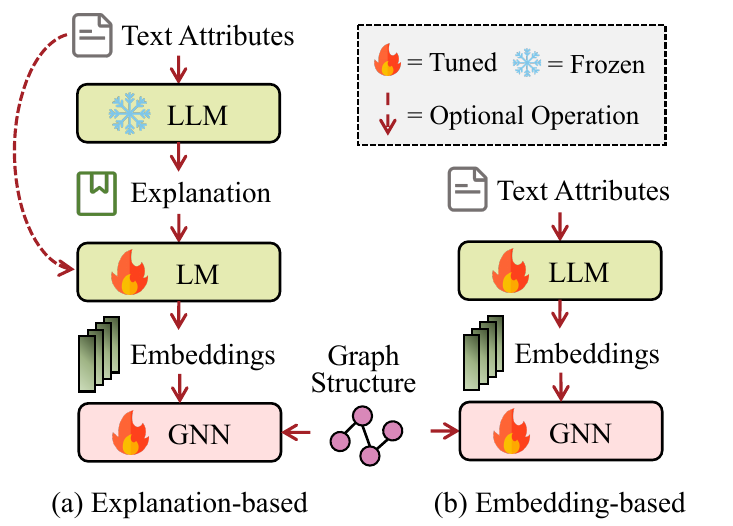}}
    \caption{The illustration of LLM-as-enhancer approaches: \textbf{a) explanation-based enhancement}, which uses LLMs to generate explanations of text attributes to enhance text embeddings; \textbf{b) Embedding-based enhancement}, which directly obtains text embeddings by LLMs as initial node embeddings.} 
    \label{fig:Enhancer_flow}
\end{figure}

To enrich the textual attributes, explanation-based enhancement approaches focus on utilizing the strong zero-shot capability of LLMs to capture higher-level information. As shown in \figref{fig:Enhancer_flow}(a), generally they prompt LLMs to generate semantically enriched additional information, such as explanations, knowledge entities, and pseudo labels. The typical pipeline is as follows:
\begin{align}
\begin{split}
    \text{Enhancement:}&\quad e_i=f_{\textsc{LLM}}(t_i, p),\ \mathbf{x}_i=f_{\textsc{LM}}(e_i, t_i),\\
    \text{Graph Learning:}&\quad\mathbf{H}=f_{\textsc{GNN}}(\mathbf{X},\mathbf{A}),
    \nonumber
\end{split}
\end{align}
where $t_i$ is the original text attributes, $p$ is the designed textual prompts, $e_i$ is the additional textual output of LLMs, $\mathbf{x}_i\in\mathbb{R}^D$ and $\mathbf{X}\in\mathbb{R}^{N\times D}$ denotes the enhanced initial node embedding of node $i$ with the dimension $D$ and embedding matrix, along with adjacency matrix $\mathbf{A}\in\mathbb{R}^{N\times N}$ to obtain node representations $\mathbf{H}\in\mathbb{R}^{N\times d}$ by GNNs, where $d$ is the dimension of representations. For instance, TAPE \citep{he2023explanations} is a pioneer work of explanation-based enhancement, which prompts LLMs to generate explanations and pseudo labels to augment textual attributes. After that, relatively small language models are fine-tuned on both original text data and explanations to encode text semantic information as initial node embeddings. \citet{chen2023exploring} explore the potential competence of LLMs in graph learning. They first compare embedding-visible LLMs with shallow embedding methods and then propose KEA to enrich the text attributes. KEA prompts LLMs to generate a list of knowledge entities along with text descriptions and encodes them by fine-tuned PLMs and deep sentence embedding models. LLM4Mol \citep{qian2023can} attempts to employ LLMs to assist in molecular property prediction. Specifically, it uses LLMs to generate semantically enriched explanations for the original SMILES and then fine-tunes a small-scale language model to conduct downstream tasks. LLMRec \citep{wei2023llmrec} aims to utilize LLMs to figure out data sparsity and data quality issues in the graph recommendation system. It reinforces user-item interaction edges and generates user/item side information by LLMs. Lastly, it employs a lightweight GNN \citep{he2020lightgcn} to encode the augmented recommendation network.

\subsection{Embedding-based Enhancement}

Refer to \figref{fig:Enhancer_flow}(b), embedding-based enhancement approaches directly utilize LLMs to output text embeddings as initial node embeddings for GNN training:
\begin{align}
    \begin{split}
    \text{Enhancement:}&\quad\mathbf{x}_i=f_{\textsc{LLM}}(t_i),\\
        \text{Graph Learning:}&\quad\mathbf{H}=f_{\textsc{GNN}}(\mathbf{X},\mathbf{A}).
        \nonumber
    \end{split}
\end{align}
This kind of approach requires the use of embedding-visible or open-source LLMs because it needs to access text embeddings straightaway and fine-tune LLMs with structural information. Many of the current advanced LLMs (e.g., GPT4 \citep{openai2023gpt4} and PaLM \citep{chowdhery2022palm}) are closed-source and only provide online services. Strict restrictions prevent researchers from accessing their parameters and output embeddings. This kind of approach mostly adopts a cascading form and utilizes structure information to assist the language model in pre-training or fine-tuning. Typically, G$\scriptstyle \text{A}$LM \citep{xie2023graph} pre-trains PLMs and GNN aggregator on a given large graph corpus to capture the information that can maximize utility towards massive applications and then fine-tunes the framework on a specific downstream application to further improve the performance.

Several works aim to generate node embeddings by incorporating structural information into the fine-tuning phase of LLMs. Representatively, GIANT \citep{chien2021node} fine-tunes the language model by a novel self-supervised learning framework, which employs XR-Transformers to solve extreme multi-label classification over link prediction. SimTeG \citep{duan2023simteg} and TouchUp-G \citep{zhu2023touchup} follow a similar way, they both fine-tune PLMs through link-prediction-like methods to help them perceive structural information. The subtle difference between them is that TouchUp-G uses negative sampling during link prediction, while SimTeG employs parameter-efficient fine-tuning to accelerate the fine-tuning process. G-Prompt \citep{huang2023prompt} introduces a graph adapter at the end of PLMs to help extract graph-aware node features. Once trained, task-specific prompts are incorporated to produce interpretable node representations for various downstream tasks. WalkLM \citep{tan2023walklm} is an unsupervised generic graph representation learning method. The first step of it is to generate attributed random walks on the graph and compose roughly meaningful textual sequences by automated textualization program. The second step is to fine-tune an LLM using textual sequences and extract representations from LLM. METERN \citep{jin2023learning} introduces relation prior tokens
to capture the relation-specific signals and uses one language encoder to model the shared knowledge across relations. LEADING \citep{xue2023efficient} effectively finetunes LLMs and transfers risk knowledge in LLM to downstream GNN model with less computation cost and memory overhead.

A recent work, OFA \citep{liu2023one}, attempts to propose a general graph learning framework, which can utilize a single graph model to conduct adaptive downstream prediction. It describes all nodes and edges using human-readable texts and encodes them from different domains into the same space by LLMs. Subsequently, the framework is adaptive to perform different tasks by inserting task-specific prompting substructures into the input graph. 

\subsection{Discussions} 

LLM-as-enhancer approaches have demonstrated superior performance on TAG, being able to effectively capture both textual and structural information. Moreover, they also exhibit strong flexibility, as GNNs and LLMs are plug-and-play, allowing them to leverage the latest techniques to address the encountered issues. Another advantage of such methods (specifically explanation-based enhancement) is that they pave the way for using closed-source LLMs to assist graph-related tasks. However, despite some papers claiming strong scalability, in fact, LLM-as-enhancer approaches entail significant overhead when dealing with large-scale datasets. Taking explanation-based approaches as an example, they need to query LLMs' APIs for $N$ times for a graph with $N$ nodes, which is indeed a substantial cost.

%% file: 5_predictor.tex
\section{LLM as Predictor}
\label{sec:predictor}

The core idea behind this category is to utilize LLMs to make predictions for a wide range of graph-related tasks, such as classifications and reasonings, within a unified generative paradigm. However, applying LLMs to graph modalities presents unique challenges, primarily because graph data often lacks straightforward transformation into sequential text, as different graphs define structures and features in different ways. In this section, we classify the models broadly into flatten-based and GNN-based predictions, depending on whether they employ GNNs to extract structural features for LLMs.

\subsection{Flatten-based Prediction}

\begin{figure}[t]
    \centering
    \resizebox{1.0\linewidth}{!}{
    \includegraphics{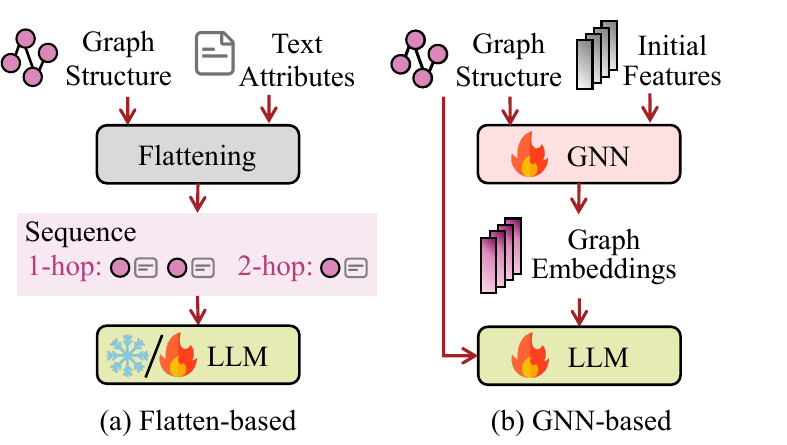}}
    \caption{The illustration of LLM-as-predictor approaches: \textbf{a) Flatten-based prediction}, which incorporates graph structure with LLMs via different flattening strategies; \textbf{b) GNN-based prediction}, utilizing GNNs to capture structural information for LLMs.} 
    \label{fig:Predictor_flow}
\end{figure}

The majority of the existing attempts that utilize LLMs as predictors employ the strategy of flattening the graph into textual descriptions, which facilitates direct processing of graph data by LLMs through text sequences. As shown in \figref{fig:Predictor_flow}(a), flatten-based prediction typically involves two steps: \textbf{(1)} utilizing a flatten function $\texttt{Flat}(\cdot)$ to transform a graph structure into a sequence of nodes or tokens $G_{seq}$, and \textbf{(2)} a parsing function $\texttt{Parse}(\cdot)$ is then applied to retrieve the predicted label from the output generated by LLMs, as illustrated below:
\begin{align}
\begin{split}
    \text{Graph Flattening:}&\quad G_{seq}=\texttt{Flat}(\mathcal{V}, \mathcal{E}, \mathcal{T}, \mathcal{J}),\\
    \text{Prediction:}&\quad \tilde{Y} = \texttt{Parse}(f_{\textsc{LLM}}(G_{seq}, p)),
    \nonumber
\end{split}
\end{align}
\noindent where $\mathcal{V}$, $\mathcal{E}$, $\mathcal{T}$, and $\mathcal{J}$ denotes the set of nodes, edges, node text attributes, and edge text attributes, respectively. $p$ indicates the instruction prompt for the current graph task and $\tilde{Y}$ is the predicted label. 

The parsing strategies of models are generally standardized.  For example, given that the output of LLMs often involves their reasoning and logic processes, particularly in the chain-of-thought (CoT) scenario, several works \citep{fatemi2023talk,zhao2023graphtext,chen2023exploring,guo2023gpt4graph,liu2023evaluating,wang2023can} utilize regular expressions to extract the predicted label from the output. Some models \citep{chen2023exploring,fatemi2023talk,wang2023can,chai2023graphllm,huang2023can} further set the decoding temperature of the LLM to $0$, in order to reduce the variance of LLM's predictions and obtain more reliable results. Another direction is to formulate graph tasks as multi-choice QA problems \citep{robinson2022leveraging} where LLMs are instructed to select the correct answer among provided choices. For instance, some works \citep{huang2023can,hu2023beyond,shi2023relm} constrain LLM's output format via giving choices and appending instructions in prompts in zero-shot setting, such as ``\textit{Do not give any reasoning or logic for your answer}''. In addition, some methods, such as GIMLET \citep{zhao2023gimlet} and InstructGLM \citep{ye2023natural}, fine-tune LLMs to directly output predicted labels, empowering them to provide accurate predictions without the need for additional parsing steps.
% + in-context learning?

Compared to parsing strategies, flattening strategies can exhibit significant variation. In the following, we organize the methods for flattening based on whether the parameters of LLMs are updated.

\subsubsection{LLM Frozen}

GPT4Graph \citep{guo2023gpt4graph} utilizes graph description languages such as GML \citep{himsolt1997gml} and GraphML \citep{brandes2013graph} to represent graphs. These languages provide standardized syntax and semantics for representing the nodes and edges within a graph. Inspired by linguistic syntax trees \citep{chiswell2007mathematical}, GraphText \citep{zhao2023graphtext} leverages graph-syntax trees to convert a graph structure to a sequence of nodes, which is then fed to LLMs for training-free graph reasoning. Furthermore, ReLM \citep{shi2023relm} uses simplified molecular input line entry system (SMILES) strings to provide one-dimensional linearizations of molecular graph structures. Graph data can be also represented through methods like adjacency matrices and adjacency lists. Several methods \citep{wang2023can,fatemi2023talk,liu2023evaluating,zhang2023llm4dyg} directly employ numerically organized node and edge lists to depict the graph data in plain text. GraphTMI \citep{das2023modality} further explores different modalities such as motif and image to integrate graph data with LLMs.

Instead, the use of natural narration to express graph structures is also making steady progress. \cite{chen2023exploring} and \cite{hu2023beyond} both integrate the structural information of citation networks into the prompts, which is achieved by explicitly representing the edge relationship through the word ``cite'' and representing the nodes using paper indexes or titles. \cite{huang2023can}, on the other hand, does not use the word ``cite'' to represent edges but instead describes the relationships via enumerating randomly selected $k$-hop neighbors of the current node. In addition, GPT4Graph \citep{guo2023gpt4graph} and \cite{chen2023exploring} imitate the aggregation behavior of GNNs and summarize the current neighbor's attributes as additional inputs, aiming to provide more structural information. It is worth noting that \cite{fatemi2023talk} investigates various methodologies to represent nodes and edges, examining a total of 11 strategies. For example, they use indexes or alphabet letters to denote nodes and apply arrows or parentheses to signify edges. 

\subsubsection{LLM Tuning}

GIMLET \citep{zhao2023gimlet} adopts distance-based position embedding to extend the capability of LLMs to perceive graph structures. When performing positional encoding of the graph, GIMLET defines the relative position of two nodes as the shortest distance between them in the graph, which has been widely utilized in the literature of graph transformers \citep{ying2021transformers}. Similar to \cite{huang2023can}, InstructGLM \citep{ye2023natural} designs a series of scalable prompts based on the maximum hop level. These prompts allow a central paper node to establish direct associations with its neighbors up to any desired hop level by utilizing the described connectivity relationships expressed in natural language.

\begin{figure*}[t]
    \centering
    \resizebox{1.0\linewidth}{!}{
    \includegraphics{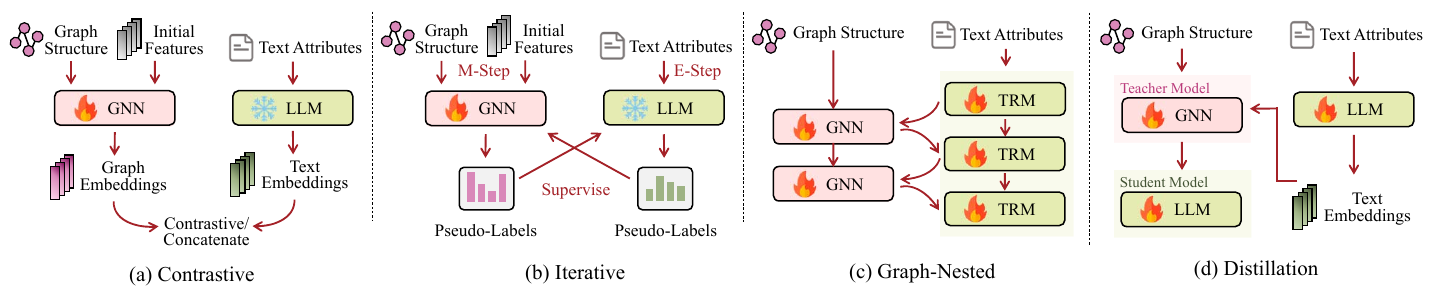}}
    \caption{The illustration of GNN-LLM-Alignment approaches: \textbf{a) Contrastive}, symmetric alignment which applies concatenation or contrastive learning to graph embeddings and text embeddings; \textbf{b) Iterative}, belongs to symmetric alignment, aiming to implement iterative interactions on embeddings of two modalities; \textbf{c) Graph-nested}, a symmetric alignment which interweaves GNNs with Transformers and \textbf{d) Distillation}, belongs to asymmetric alignment, which uses GNN as a teacher to train language models to be graph-aware.} 
    \label{fig:Alignment_flow}
\end{figure*}

\subsection{GNN-based Prediction}

GNNs have demonstrated impressive capabilities in understanding graph structures through recursive information exchange and aggregation among nodes. As illustrated in \figref{fig:Predictor_flow}(b), in contrast to flatten-based prediction, which converts graph data into textual descriptions as inputs to LLMs, GNN-based prediction leverages the advantages of GNNs to incorporate inherent structural characteristics and dependencies present in graph data with LLMs, allowing LLMs to be structure-aware as follows: 
\begin{align}
\begin{split}
    \text{Graph Learning:}&\quad\mathbf{H}=f_{\textsc{GNN}}(\mathbf{X},\mathbf{A}), \\
    \text{Prediction:}&\quad \tilde{Y} = \texttt{Parse}(f_{\textsc{LLM}}(\mathbf{H}, p)),
    \nonumber
\end{split}
\end{align}
\noindent where $\mathbf{X}$ denotes the node embedding matrix, $\mathbf{A}$ is the adjacency matrix, and $\mathbf{H}$ denotes the structure-aware embeddings associated with the graph. GNN-based prediction also relies on a parser to extract the output from LLMs. However, integrating GNN representations into LLMs often requires tuning, making it easier to standardize the prediction format of LLMs by providing desirable outputs during training.

Various strategies have been proposed to fuse the structural patterns learned by GNNs and the contextual information captured by LLMs. For instance, GIT-Mol \citep{liu2023git} and MolCA \citep{liu2023molca} both implement BLIP-2's Q-Former \citep{li2023blip} as the cross-modal projector to map the graph encoder's output to the LLM's input text space. Multiple objectives with different attention masking strategies are employed for effective graph-text interactions. GraphLLM \citep{chai2023graphllm} derives the graph-enhanced prefix by applying a linear projection to the graph representation during prefix tuning, allowing the LLM to synergize with the graph transformer to incorporate structural information crucial to graph reasoning. Additionally, both GraphGPT \citep{tang2023graphgpt} and InstructMol \citep{cao2023instructmol} employ a simple linear layer as the lightweight alignment projector to map the encoded graph representation to some graph tokens, while the LLM excels at aligning these tokens with diverse text information. DGTL \citep{qin2023disentangled} injects the disentangled graph embeddings directly into each layer of the LLM, highlighting different aspects of the graph’s topology and semantics.

\subsection{Discussions}

Utilizing LLMs directly as predictors shows superiority in processing textual attributes of graphs, especially achieving remarkable zero-shot performance compared with traditional GNNs. The ultimate goal is to develop and refine methods for encoding graph-structured information into a format that LLMs can comprehend and manipulate effectively and efficiently. Flatten-based prediction may have an advantage in terms of effectiveness, while GNN-based prediction tends to be more efficient. In flatten-based prediction, the input length limitation of LLMs restricts each node's access to only its neighbors within a few hops, making it challenging to capture long-range dependencies. Additionally, without the involvement of GNNs, inherent issues of GNNs such as heterophily cannot be addressed. On the other hand, for GNN-based prediction, training an additional GNN module and inserting it into LLMs for joint training is challenging due to the problem of vanishing gradients in the early layers of deep transformers \citep{zhao2023gimlet,qin2023disentangled}.

%% file: 6_align.tex
\section{GNN-LLM Alignment}

Aligning the embedding spaces of GNNs and LLMs is an effective way to integrate the graph modality with the text modality. GNN-LLM alignment ensures that each encoder's unique functionalities are preserved while coordinating their embedding spaces at a specific stage. In this section, we summarize the techniques for aligning GNNs and LLMs, which can be classified as symmetric or asymmetric, depending on whether equal emphasis is placed on both GNNs and LLMs or if one modality is prioritized over the other.

\subsection{Symmetric}

Symmetric alignment refers to the equal treatment of the graph and text modalities during the alignment process. These approaches ensure that the encoders of both modalities achieve comparable performance in their respective applications.

A typical symmetric alignment architecture, illustrated in \figref{fig:Alignment_flow}(a), adopts a two-tower style, employing separate encoders to individually encode the graph and text. Notice that during the alignment process, both modalities interact only once. Some methods, like SAFER \citep{chandra2020graph}, utilize simple concatenation on these separate embeddings.
However, this approach falls short in achieving a seamless fusion of structural and textual information, resulting in a loosely coupled integration of the two modalities. Consequently, the majority of two-tower style models utilize contrastive learning techniques to facilitate alignment, akin to CLIP \citep{radford2021learning} for aligning visual and language modalities. In general, the methods encompass two steps. The first step is feature extraction, where graph representations and text representations are obtained. The second step involves the use of a contrastive learning process with a modified InfoNCE loss \citep{oord2018representation} depicted with the following equation: 
\begin{gather}
    \ell(\mathbf{g}_i,\mathbf{t}_i)=-\log\frac{e^{s(\mathbf{g}_i,\mathbf{t}_i)/\tau}}{\sum_{k=1}^{|\mathcal{G}|}e^{s(\mathbf{g}_i,\mathbf{t}_k)/\tau}},\nonumber\\
    \mathcal{L}_{\text{InfoNCE}}=\frac{1}{2|\mathcal{G}|}\sum_{i=1}^{|\mathcal{G}|}\Big(\ell(\mathbf{g}_i,\mathbf{t}_i)+\ell(\mathbf{t}_i,\mathbf{g}_i)\Big),
    \nonumber
\end{gather}
where $\mathbf{g}$ represents the representation of a specific graph, while $\mathbf{t}$ denotes the representation of the corresponding text of the graph. $s(\cdot,\cdot)$ denotes the score function that assigns high values to the positive pair, and low values to negative pairs. $\tau$ is a temperature parameter and $|\mathcal{G}|$ denotes the number of graphs in the training dataset. Parameters of both encoders are updated via backpropagation based on the contrastive loss.

\input{tables/models}

Text2Mol \citep{edwards2021text2mol} proposes a cross-modal attention mechanism to achieve early fusion of graph and textual embeddings. Implemented through a transformer decoder, Text2Mol uses the LLM's output as a source sequence and the GNN's output as a target sequence. This setup allows the attention mechanism to learn multimodal association rules. The decoder's output is then utilized for contrastive learning, paired with the processed outputs from the GNN.

MoMu \citep{su2022molecular}, MoleculeSTM \citep{liu2022multi}, ConGraT \citep{brannon2023congrat}, and RLMRec \citep{ren2023representation} share a similar framework, which adopts paired graph embeddings and text embeddings to implement contrastive learning, but there are still differences in detail. Both MoMu and MoleculeSTM gather molecules from PubChem \citep{wang2009pubchem}. The former retrieves related texts from published scientific papers, while the latter utilizes the corresponding descriptions of the molecules. ConGraT expands this architecture beyond the molecular domain. It has validated this graph-text paired contrastive learning method on social, knowledge, and citation networks. RLMRec proposes to align the semantic space of LLMs with the representation space of collaborative relational signals (which indicate user-item interactions) in recommendation systems through a contrastive modeling.

Several studies such as G2P2 \citep{wen2023prompt} and G$\scriptstyle \text{RENADE}$ \citep{li2023grenade} have further advanced the use of contrastive learning. Specifically, G2P2 enhances the granularity of contrastive learning and introduces prompts during the fine-tuning stage. It employs contrastive learning at three levels during the pre-training stage: node-text, text-text summary, and node-node summary, thereby strengthening the alignment between text and graph representations. Prompts are utilized in downstream tasks, demonstrating strong performance in few-shot and zero-shot text classification and node classification tasks. On the other hand, G$\scriptstyle \text{RENADE}$ is optimized by integrating graph-centric contrastive learning with dual-level graph-centric knowledge alignment, which includes both node-level and neighborhood-level alignment.

Contrary to previous methods, the iterative alignment approach, depicted in \figref{fig:Alignment_flow}(b), treats both modalities equally but distinguishes itself in the training process by allowing for iterative interaction between the modalities. For example, GLEM \citep{zhao2022learning} employs the Expectation-Maximization (EM) framework, where one encoder iteratively generates pseudo-labels for the other encoder, allowing them to align their representation spaces.

\subsection{Asymmetric}

While symmetric alignment aims to give equal emphasis to both modalities, asymmetric alignment focuses on allowing one modality to assist or enhance the other. In current studies, the predominant approach involves leveraging the capabilities of GNNs to process structural information to reinforce LLMs. These studies can be categorized into two types: graph-nested transformer and graph-aware distillation.

The graph-nested transformer, as exemplified by Graphformer \citep{yang2021graphformers} in \figref{fig:Alignment_flow}(c), demonstrates asymmetric alignment through the integration of GNNs into each transformer layer. Within each layer of the LLM, the node embedding is obtained from the first token-level embedding, which corresponds to the [\texttt{CLS}] token. The process involves gathering embeddings from all relevant nodes and applying them to a graph transformer. The output is then concatenated with the input embeddings and passed on to the next layer of the LLM. Patton \citep{jin2023patton} extends GraphFormer by proposing two pre-training strategies, i.e., network-contextualized masked language modeling and masked node prediction, specifically for text-rich graphs. Its strong performance is shown in various downstream tasks, including classification, retrieval, reranking, and link prediction.

Additionally, G$\scriptstyle \text{RA}$D \citep{mavromatis2023train} employs graph-aware distillation for aligning two modalities, depicted in \figref{fig:Alignment_flow}(d). It utilizes a GNN as a teacher model to generate soft labels for an LLM, facilitating the transfer of aggregated information. Moreover, since the LLMs share parameters, the GNN can benefit from improved textual encodings after the updates to the LLMs' parameters. Through iterative updates, a graph-aware LLM is developed, resulting in enhanced scalability in inference due to the absence of the GNN. Similar to G$\scriptstyle \text{RA}$D, THLM \citep{zou2023pretraining} employs a heterogeneous GNN to enhance LLMs with multi-order topology learning capabilities. It involves pretraining a LLM alongside an auxiliary GNN through two distinct strategies. The first strategy focuses on predicting whether a node is part of the context graph of a target node. The second strategy utilizes a Masked Language Modeling task, which aids in developing a robust language comprehension by the LLM. After the pretraining process, the auxiliary GNN is discarded and the LLM is fine-tuned for downstream tasks.

\subsection{Discussions}

To align GNNs and LLMs, symmetric alignments treat each modality equally, with the objective of enhancing GNNs and LLMs simultaneously. This leads to encoders that can effectively handle tasks involving both modalities, leveraging their individual encoding strengths to improve modality-specific representations. In addition, asymmetric methods enhance LLMs by inserting graph encoders into transformers or directly using GNNs as teachers. However, alignment techniques face challenges when dealing with data scarcity. In particular, only a few graph datasets (i.e., molecular datasets) contain native graph-text pairs, limiting the applicability of these methods.

%% file: tables/models.tex
\begin{table*}[t]
\resizebox{\linewidth}{!}{
\begin{tabular}{llllcccllc}
\toprule
   & \textbf{Model} & \textbf{GNN} & \textbf{LLM} & \textbf{Predictor}  & \textbf{Fine-tuning} & \textbf{Prompting}  & \textbf{Domain}  & \textbf{Task} & \textbf{Code} \\ \midrule
 
\multirow{13}{*}{\rotatebox{90}{LLM as Enhancer}}    & GIANT \citep{chien2021node}   & SAGE, RevGAT, etc.    & BERT  & GNN     & \ding{55}      & \ding{55}          & Citation, Co-purchase          & Node     & \href{https://github.com/amzn/pecos/tree/mainline/examples/giant-xrt}{Link}     \\
& G$\scriptstyle \text{A}$LM \citep{xie2023graph}  &  RGCN, RGAT  &    BERT    &   GNN      &  \ding{51}    &  \ding{55}      &  E-Commerce, Recommendation  &   Node, Link   & -\\
& TAPE \citep{he2023explanations}  &   RevGAT  & ChatGPT       & GNN        & \ding{55} & \ding{51}  &   Citation     & Node     & \href{https://github.com/XiaoxinHe/TAPE}{Link}     \\
& Chen et al. \citep{chen2023exploring}   & GCN, GAT   & ChatGPT  & GNN          & \ding{55}          & \ding{51}          & Citation, Co-purchase     & Node    & - \\
 & LLM4Mol \citep{qian2023can}   & -   & ChatGPT       & LM       & \ding{55}     & \ding{55}       & Molecular   & Graph     & \href{https://github.com/ChnQ/LLM4Mol}{Link}     \\
 & SimTeG \citep{duan2023simteg}   & SAGE, RevGAT, SEAL   & allMiniLM-L6-v2, etc.    & GNN         & \ding{51}$^\heartsuit$      & \ding{55}    & Citation, Co-purchase          & Node, Link     & \href{https://github.com/vermouthdky/SimTeG}{Link}     \\
 & G-Prompt \citep{huang2023prompt}  & SAGE, RevGAT   & RoBERTa-Large       & GNN      & \ding{51}     & \ding{51}       & Citation, Social   & Node     & -     \\
 & TouchUp-G \citep{zhu2023touchup}   & SAGE, MB-GCN, etc.    & BERT   & GNN   & \ding{51}       & \ding{55}          & Citation, Co-purchase, Recommendation          & Node, Link     & -     \\
 & OFA \citep{liu2023one}   & R-GCN   & Sentence-BERT       & GNN     & \ding{55}          &   \ding{51}     & Citation, Web link, Knowledge, Molecular   & Node, Link, Graph     & \href{https://github.com/LechengKong/OneForAll}{Link}     \\
& LLMRec \citep{wei2023llmrec}  & LightGCN   & ChatGPT      & GNN      & \ding{55}     & \ding{51}       & Recommendation   & Recommendation     & \href{https://github.com/HKUDS/LLMRec}{Link}     \\
& WalkLM \citep{tan2023walklm}  & -   & DistilRoBERTa      & MLP      & \ding{51}     & \ding{55}       & Knowledge   & Node, Link     & \href{https://github.com/Melinda315/WalkLM}{Link}     \\
& METERN \citep{jin2023learning}  & -   & BERT      & LM      & \ding{51}     & \ding{55}       & Citation, E-Commerce   & Node     & -     \\
& LEADING \citep{xue2023efficient}  & GCN, GAT   & BERT      & GNN      & \ding{51}     & \ding{55}       & Citation   & Node     & -     \\
                                   \midrule

\multirow{19}{*}{\rotatebox{90}{LLM as Predictor}}  & NLGraph \citep{wang2023can} & -   &  Text-davinci-003 &   LLM      &   \ding{55}  & \ding{51}    &  -   &  Reasoning    &  \href{https://github.com/Arthur-Heng/NLGraph}{Link}   \\

 &  GPT4Graph \citep{guo2023gpt4graph}  &  -  &   Text-davinci-003     &    LLM  &   \ding{55}    &   \ding{51}     &  -  &   Reasoning, Node, Graph   & \href{https://anonymous.4open.science/r/GPT4Graph}{Link}     \\
     
 & GIMLET \citep{zhao2023gimlet} &      -    &   T5   & LLM           &  \ding{51}/\ding{55}    &   \ding{51}    &     Molecular  &  Graph    &   \href{https://github.com/zhao-ht/GIMLET}{Link}   \\ 

&  Chen et al. \citep{chen2023exploring}  &  -  &   ChatGPT     &   LLM    &   \ding{55}   &   \ding{51}     &  Citation  &   Node   &  \href{https://github.com/CurryTang/Graph-LLM}{Link}    \\

& GIT-Mol  \citep{liu2023git}  &  GIN  &   MolT5    & LLM     &  \ding{51}$^\heartsuit$    &   \ding{51}     &  Molecular  &  Graph, Captioning    &   -   \\

&  InstructGLM \citep{ye2023natural} &  -  &   FLAN-T5/LLaMA-v1     &    LLM  &   \ding{51}$^\heartsuit$   &   \ding{51}     &  Citation  &  Node    &   \href{https://github.com/agiresearch/InstructGLM}{Link}  \\

&  Liu et al. \citep{liu2023evaluating}  &  -  &  GPT-4, etc.  &   LLM  &    \ding{55}   &  \ding{51}      &  -  &  Reasoning    & \href{https://github.com/Ayame1006/LLMtoGraph}{Link}     \\

&  Huang et al. \citep{huang2023can} &  -  &  ChatGPT      &   LLM     &  \ding{55}   &   \ding{51}    &  Citation, Co-purchase  &  Node    &  \href{https://github.com/TRAIS-Lab/LLM-Structured-Data}{Link}    \\

& GraphText \citep{zhao2023graphtext}  &  -  &   ChatGPT/GPT-4   &  LLM     &  \ding{55}    &   \ding{51}     &  Citation, Web link  &   Node   &  -   \\

&  Fatemi et al. \citep{fatemi2023talk}  &  -  &   PaLM/PaLM 2     &    LLM   &   \ding{55}   &   \ding{51}     &   - &  Reasoning    &    -  \\

& GraphLLM \citep{chai2023graphllm}  &  Graph Transformer   &   LLaMA-v2     &  LLM    &   \ding{51}$^\heartsuit$   &   \ding{51}     &   - &  Reasoning    &   \href{https://github.com/mistyreed63849/Graph-LLM}{Link}   \\

& Hu et al. \citep{hu2023beyond}  &  -  &  ChatGPT/GPT-4   &      LLM   &   \ding{55}    &    \ding{51}     &  Citation, Knowledge, Social  &   Node, Link, Graph   &    -  \\

& MolCA \citep{liu2023molca}  &  GINE  &    Galactica/MolT5   &   LLM    &  \ding{51}$^\heartsuit$    &  \ding{51}      &  Molecular  &  Graph, Retrieval, Captioning    &   \href{https://github.com/acharkq/MolCA}{Link}   \\

&  GraphGPT \citep{tang2023graphgpt}  &  Graph Transformer  &  Vicuna    &  LLM     &   \ding{51}$^\heartsuit$   &   \ding{51}     &  Citation  &  Node    &  \href{https://github.com/HKUDS/GraphGPT}{Link}    \\

&  ReLM \citep{shi2023relm}  &   TAG, GCN &  Vicuna/ChatGPT    & LLM    &  \ding{55}    &   \ding{51}     &  Molecular  &  Reaction Prediction    &    \href{https://github.com/syr-cn/ReLM}{Link}   \\

& LLM4DyG \citep{zhang2023llm4dyg} &  -  &  Vicuna/LLaMA-v2/ChatGPT    &  LLM   &  \ding{55}   &    \ding{51}   &  -  & Reasoning    & -      \\

& DGTL \citep{qin2023disentangled} & Disentangled GNN   & LLaMA-v2    & LLM    & \ding{51}    & \ding{51}      & Citation, E-Commerce  & Node    & -      \\

& GraphTMI \citep{das2023modality} & -   & GPT-4/GPT-4V    & LLM    & \ding{55}    & \ding{51}      & Citation & Node    & -      \\

& InstructMol \citep{cao2023instructmol} & GIN   & Vicuna    & LLM    & \ding{51}$^\heartsuit$    & \ding{51}      & Molecular & Graph, Captioning   & \href{https://github.com/IDEA-XL/InstructMol}{Link}     \\

 \midrule
\multirow{13}{*}{\rotatebox{90}{GNN-LLM Alignment}} 
&  SAFER\citep{chandra2020graph}  & GCN, GAT, etc.   &   RoBERTa        &  Linear      &  \ding{51}      &  \ding{55}    &    News  &   Node   &   \href{https://github.com/shaanchandra/SAFER }{Link}    \\
& GraphFormers \citep{yang2021graphformers}  & Graph Transformer &   UniLM        &  LLM    &    \ding{51}   & \ding{55}   &    Citation, E-Commerce, Knowledge   &  Link    &   \href{https://github.com/microsoft/GraphFormers}{Link}    \\
& Text2Mol\citep{edwards2021text2mol}  & GCN   &      SciBERT     &   GNN/LLM   &    \ding{51}   & \ding{55}   &  Molecular     & Retrieval      &   \href{https://github.com/cnedwards/text2mol}{Link}    \\
&  MoMu \citep{su2022molecular}  & GIN   &   BERT        &  GNN/LLM       &  \ding{51}    &  \ding{55}    &    Molecular  &   Graph, Retrieval   &   \href{https://github.com/BingSu12/MoMu }{Link}    \\
&  MoleculeSTM \citep{liu2022multi}    &   GIN   &  BERT    &       GNN/LLM     &    \ding{51}       &    \ding{55}       &  Molecular    &  Graph, Retrieval & \href{https://github.com/chao1224/MoleculeSTM}{Link}    \\ 
& GLEM \citep{zhao2022learning}  & SAGE, RevGAT, etc.   &   DeBERTa     &   GNN/LLM    &   \ding{51}     &   \ding{55}     &  Citation, Co-purchase  &  Node&\href{https://github.com/AndyJZhao/GLEM}{Link}        \\
& G$\scriptstyle \text{RA}$D \citep{mavromatis2023train}  & SAGE   &      SciBERT/DistilBERT  &   LLM    &    \ding{51}     &   \ding{55}     &  Citation, Co-purchase  &  Node&\href{https://github.com/cmavro/GRAD}{Link}        \\
& G2P2 \citep{wen2023prompt}  &  GCN  &   Transformer     &   GNN/LLM   &    \ding{51}   &     \ding{51}   &  Citation, Recommendation   & Node     &   \href{https://github.com/WenZhihao666/G2P2-conditional}{Link}   \\
& Patton \citep{jin2023patton}  & Graph Transformer  &    BERT/SciBERT     &    Linear/LLM   &  \ding{51}     &   \ding{55}     &  Citation, E-Commerce  &  Node, Link, Retrieval, Reranking    &   \href{https://github.com/PeterGriffinJin/Patton}{Link}   \\
& ConGraT \citep{brannon2023congrat}  &  GAT  &  all-mpnet-base-v2/DistilGPT2      &    GNN/LLM    &  \ding{51}     &   \ding{55}     & Citation, Knowledge, Social   &  Node, Link    &   \href{https://github.com/wwbrannon/congrat}{Link}   \\
% & G$\scriptstyle \text{A}$LM \citep{xie2023graph}  &  RGCN, RGAT  &    BERT    &   GNN     &   \ding{51}   &  \ding{51}    &  \ding{55}      &  E-Commerce, Recommendation  &   Node, Link   & -\\
& THLM \citep{zou2023pretraining}  &  R-HGNN  &   BERT   &  LLM      & \ding{51}     &  \ding{55}    & Academic, Recommendation, Patent  &   Node, Link   & \href{https://github.com/Hope-Rita/THLM}{Link}   \\
& G$\scriptstyle \text{RENADE}$ \citep{li2023grenade}  & SAGE, RevGAT-KD, etc.   &   BERT     &   GNN/MLP    &    \ding{51}   &     \ding{55}    & Citation, Co-purchase      &   Node, Link   &  \href{https://github.com/bigheiniu/GRENADE}{Link}    \\ 
& RLMRec \citep{ren2023representation}  & GCCF, LightGCN, etc.   &   ChatGPT, text-embedding-ada-002    &   GNN/LLM    &    \ding{51}   &     \ding{55}    & Recommendation     &   Node  &  \href{https://github.com/HKUDS/RLMRec}{Link}    \\ 
\midrule

\multirow{3}{*}{\rotatebox{90}{Others}} 
& LLM-GNN \citep{chen2023label}  & GCN, SAGE   & ChatGPT       & GNN       & \ding{55}     & \ding{51}       & Citation, Co-purchase   & Node     & \href{https://github.com/CurryTang/LLMGNN}{Link}     \\

& GPT4GNAS \citep{wang2023graph}  &  GCN, GIN, etc.  &   GPT-4  &   GNN    &   \ding{55}   &   \ding{51}    & Citation     &  Node   &   -  \\

& ENG \citep{yu2023empower}  & GCN, GAT   & ChatGPT       & GNN       & \ding{55}     & \ding{51}       & Citation   & Node     & -     \\
                      \bottomrule
\end{tabular}}
\caption{A summary of models that leverage LLMs to assist graph-related tasks in literature, ordered by their release time. \textbf{Fine-tuning} denotes whether it is necessary to fine-tune the parameters of LLMs, and $\heartsuit$ indicates that models employ parameter-efficient fine-tuning (PEFT) strategies, such as LoRA and prefix tuning. \textbf{Prompting} indicates the use of text-formatted prompts in LLMs, done manually or automatically. Acronyms in \textbf{Task}: Node refers to node-level tasks; Link refers to link-level tasks; Graph refers to graph-level tasks; Reasoning refers to Graph Reasoning; Retrieval refers to Graph-Text Retrieval; Captioning refers to Graph Captioning.}
\label{tab:summarizations}
\end{table*}

%% file: 7_future.tex
\section{Future Directions}

\tabref{tab:summarizations} summarizes the models that leverage LLMs to assist graph-related tasks according to the proposed taxonomy. Based on the above review and analysis, we believe that there is still much space for further enhancement in this field. In this section, we discuss the remaining limitations of leveraging LLM's ability to comprehend graph data and list some directions for further exploration in subsequent research.

\paratitle{Dealing with non-TAG.}
Utilizing LLMs to assist learning on text-attributed graphs has already demonstrated excellent performance. However, graph-structured data is ubiquitous in real-world scenarios, and a great deal of it lacks rich textual information. For example, in a traffic network (e.g., PeMS03 \citep{song2020spatial}), each node represents an operational sensor, while in a superpixel graph (e.g., PascalVOC-SP \citep{dwivedi2022long}), each node represents a superpixel. These datasets do not have attached text attributes on each node, and it is also challenging to describe the semantic meaning of each node using human-understandable language. Although OFA \citep{liu2023one} proposes to describe all nodes and edges using human-understandable texts and embed the texts into the same space by LLMs, it may not be applicable to all domains (e.g., superpixel graph), and its performance may be suboptimal in certain domains and datasets. Exploring how to leverage the powerful generalization capabilities of LLMs to help in constructing graph foundation models is a valuable research direction.

\paratitle{Dealing with data leakage.}
Data leakage in LLMs has become a focal point of discussion \citep{aiyappa2023can}. Given that LLMs undergo pre-training on extensive text corpora, it’s likely that LLMs may have seen and memorized at least part of the test data of the common benchmark datasets, especially for citation networks. This undermines the reliability of current studies that rely on earlier benchmark datasets. In addition, \cite{chen2023exploring} proves that specific prompts could potentially enhance the ``activation'' of LLMs’ corresponding memory, thereby influencing the evaluation. Both \cite{huang2023can} and \cite{he2023explanations} have tried to avoid the data leakage issue by collecting a new citation dataset, ensuring that the test papers are sampled from time periods post the data cut-off of ChatGPT. However, they still remain limited to the citation domain and the impact of graph structures in their datasets is not significant. Hence, it’s crucial to reconsider the methods employed to accurately evaluate the performance of LLMs on graph-related tasks. A fair, systematic, and comprehensive benchmark is also needed.

\paratitle{Improving transferability.}
Transferability has always been a challenging problem in the graph domain \citep{jiang2022transferability}. The transferability of learned knowledge from one dataset to another, or from one domain to another, is not straightforward due to the unique characteristics and structures of individual graphs. Graphs can vary significantly in terms of size, connectivity, node types, edge types, and overall topology, making it difficult to directly transfer knowledge between them. While LLMs have demonstrated promising zero/few-shot abilities in language tasks due to their extensive pre-training on vast amounts of corpora, the exploration of utilizing the knowledge embedded within LLMs to enhance the transferability of graph-related tasks has been relatively limited. OFA \citep{liu2023one} attempts a unified way to perform cross-domain on graphs by describing all nodes and edges as human-readable texts and embedding the texts from different domains into the same embedding space with a single LLM. The topic of improving transferability is still worth investigating.

\paratitle{Improving explainability.} 
Explainability, also known as interpretability, denotes the ability to explain or present the behavior of models in human-understandable terms \citep{zhao2023explainability}. LLMs exhibit improved explainability compared to GNNs when handling graph-related tasks, primarily due to the reasoning and explaining ability of LLMs to produce user-friendly explanations for graph reasoning, including generating additional explanations as enhancers discussed in \secref{sec:enhancer} and offering reasoning processes as predictors discussed in \secref{sec:predictor}. Several studies have examined explaining techniques within the prompting paradigm, such as in-context learning \citep{radford2021learning} and chain-of-thought \citep{wei2022chain}, which involve feeding a sequence of demonstrations and prompts to the LLM to steer its generation in a particular direction and have it explain its reasoning. Further explorations should be conducted to enhance explainability.

\paratitle{Improving efficiency.}
While LLMs have demonstrated their effectiveness in learning on graphs, they may face inefficiencies in terms of time and space, particularly compared to dedicated graph learning models such as GNNs that inherently process graph structures. This is especially obvious when LLMs rely on sequential graph descriptions for predictions discussed in \secref{sec:predictor}. For example, while accessing LLMs through APIs (i.e., ChatGPT and GPT-4), the billing model incurs high costs for processing large-scale graphs. Additionally, both training and inference for locally deployed open-source LLMs require significant time consumption and substantial hardware resources. Existing studies \citep{duan2023simteg,liu2023git,ye2023natural,chai2023graphllm,liu2023molca,tang2023graphgpt} have tried to enable LLMs' efficient adaption via adopting parameter-efficient fine-tuning strategies, such as LoRA \citep{hu2021lora} and prefix tuning \citep{li2021prefix}. We believe that more efficient methods may unlock more power of applying LLMs on graph-related tasks with limited computational resources.

\paratitle{Analysis and improvement of expressive ability.}
Despite the recent achievements of LLMs in graph-related tasks, their theoretical expressive power remains largely unexplored. It is widely acknowledged that standard message-passing neural networks are as expressive as the 1-Weisfeiler-Lehman (WL) test, meaning that they fail to distinguish non-isomorphic graphs under 1-hop aggregation \citep{xu2018powerful}. Therefore, two fundamental questions arise: How effectively do LLMs understand graph structures? Can their expressive ability surpass those of GNNs or the WL-test? Besides, permutation equivariance is an intriguing property of typical GNNs, which is significant in geometric graph learning \citep{han2022geometrically}. Exploring how to endow LLMs with this property is also an interesting direction.

\paratitle{LLMs as agent.}
In the current integration of graphs with LLMs, LLMs often play the role of enhancers, predictors, and alignment components. However, in more complex scenarios, such applications may not fully unlock the potential of LLMs. Recent research has explored new roles for LLMs as agents, such as generative agents \citep{park2023generative} and domain-specific agents \citep{bran2023chemcrow}. In an LLM-powered agent system, LLMs function as the agent's brain, supported by essential components like planning, memory, and tool using \citep{weng2023prompt}. In complex graph-related scenarios, such as recommendation systems and knowledge discovery, treating LLMs as agents to first decompose tasks into multiple subtasks, and then identifying the most suitable tools (e.g., GNNs) for each subtask may potentially yield enhanced performance. Furthermore, employing LLMs as agents holds promise for constructing a powerful and highly generalizable solver for graph-related tasks.

%% file: 8_con.tex
\section{Conclusion}

The application of LLMs to graph-related tasks has emerged as a prominent area of research in recent years. In this survey, we aim to provide an in-depth overview of existing strategies for adapting LLMs to graphs. Firstly, we introduce a novel taxonomy that categorizes techniques involving both graph and text modalities into three categories based on the different roles played by LLMs, i.e., enhancer, predictor, and alignment component. Secondly, we systematically review the representative studies according to the taxonomy. Finally, we discuss some limitations and highlight several future research directions. Through this comprehensive review, we aspire to shed light on the advancements and challenges in the field of graph learning with LLMs, thereby encouraging further enhancements in this domain.